\documentclass[11pt,a4paper]{article}
\usepackage{authblk}
\usepackage[hyperref]{acl2017}
\usepackage{times}
\usepackage{latexsym}
\usepackage{color}
\usepackage{url}
\usepackage[normalem]{ulem}
\usepackage{framed}

\usepackage{etoolbox}
\makeatletter
\patchcmd\@combinedblfloats{\box\@outputbox}{\unvbox\@outputbox}{}{%
  \errmessage{\noexpand\@combinedblfloats could not be patched}%
}%
\makeatother

\aclfinalcopy % Uncomment this line for the final submission
\def\aclpaperid{50} %  Enter the acl Paper ID here

\usepackage{tabularx}   
\usepackage{booktabs}       % professional-quality tables

\usepackage[
subtle,% don't mess around too much
]{savetrees}

\usepackage{acronym}
\acrodef{STFT}{Short-time Fourier transform}
\acrodef{ASR}{Automated speech recognition}
\acrodef{SDA}{stacked denoising auto-encoder}
\acrodef{CAE}{convolutional auto-encoder}
\acrodef{MLP}{multilayer perceptron}
\acrodef{DBN}{deep belief net}
\acrodef{HMM}{hidden Markov model}
\acrodef{SGD}{stochastic gradient descent}
\acrodef{CNN}{convolutional neural network}
\acrodef{RNN}{recurrent neural network}
\acrodef{ANN}{artificial neural network}
\acrodef{SVM}{support vector machine}
\acrodef{PCA}{principle component analysis}
\acrodef{MSRE}{mean square reconstruction error}
\acrodef{ASG}{AutoSegCriterion}
\acrodef{CTC}{Connectionist Temporal Classification}
\acrodef{BAS}{Bavarian Archive for Speech Signals}
\acrodef{DNN}{deep neural network}
\acrodef{WER}{word error rate}
\acrodef{LER}{letter error rate}

\newcommand{\anonymize}[2]{\textcolor{black}{#1}}		% final version
\usepackage{graphicx}
\graphicspath{ {figs/} }

\title{Transfer Learning for Speech Recognition on a Budget}

\author[1]{Julius Kunze}
\author[1]{Louis Kirsch}
\author[2]{Ilia Kurenkov}
\author[2]{Andreas Krug}
\author[2]{\authorcr Jens Johannsmeier}
\author[2]{Sebastian Stober}

\affil[1]{Hasso Plattner Institute, Potsdam, Germany \authorcr
{\tt juliuskunze@gmail.com, mail@louiskirsch.com}}
\affil[2]{University of Potsdam, Potsdam, Germany \authorcr
{\tt \{kurenkov,ankrug,johannsmeier,sstober\}@uni-potsdam.de}}

\date{}

\begin{document}
\maketitle
\begin{abstract}
End-to-end training of automated speech recognition (ASR) systems requires massive data and compute resources.
We explore transfer learning based on model adaptation as an approach for training ASR models under constrained GPU memory, throughput and training data. 
We conduct several systematic experiments adapting a Wav2Letter convolutional neural network originally trained for English ASR to the German language.
We show that this technique allows faster training on consumer-grade resources while requiring less training data in order to achieve the same accuracy, thereby lowering the cost of training ASR models in other languages. 
Model introspection revealed that small adaptations to the network's weights were sufficient for good performance, especially for inner layers.
\end{abstract}
% Reset acronyms after abstract
\acresetall

\sloppy

\section{Introduction}\label{sec:intro}

\ac{ASR} is the task of translating spoken language to text in real-time.
Recently, end-to-end deep learning approaches have surpassed previously predominant solutions based on Hidden Markov Models.
In an influential paper, \citet{DBLP:journals/corr/AmodeiABCCCCCCD15} used \acp{CNN} and \acp{RNN} to redefine the state of the art.
However, \citet{DBLP:journals/corr/AmodeiABCCCCCCD15} also highlighted the shortcomings of the deep learning approach.
Performing forward and backward propagation on complex deep networks in a reasonable amount of time requires expensive specialized hardware.
Additionally, in order to set the large number of parameters of a deep network properly, one needs to train on large amounts of audio recordings. 
Most of the time, the recordings need to be transcribed by hand.
Such data in adequate quantities is currently available for few languages other than English.

We propose an approach combining two methodologies to address these shortcomings.
Firstly, we use a simpler model with a lower resource footprint.
Secondly, we apply a technique called \emph{transfer learning} to significantly reduce the amount of non-English training data needed to achieve competitive accuracy in an \ac{ASR} task.
We investigate the efficacy of this approach on the specific example of adapting a CNN-based end-to-end model originally trained on English to recognize German speech.
In particular, we freeze the parameters of its lower layers while retraining the upper layers on a German corpus which is smaller than its English counterpart.

We expect this approach to yield the following three improvements.
Taking advantage of the representation learned by the English model will lead to shorter training times compared to training from scratch.
Relatedly, the model trained using transfer learning requires less data for an equivalent score than a German-only model.
Finally, the more layers we freeze the fewer layers we need to back-propagate through during training.
Thus we expect to see a decrease in GPU memory usage since we do not have to maintain gradients for all layers.

This paper is structured as follows.
\autoref{sec:related-work} gives an overview of other transfer learning approaches to \ac{ASR} tasks.
Details about our implementation of the Wav2Letter model and how we trained it can be found in \autoref{sec:model-architecture}.
The data we used and how we preprocessed it is described in \autoref{sec:datasets}.
After a short introduction of the performed experiments in \autoref{sec:experiments} we present and discuss the results in \autoref{sec:results} followed by a conclusion in \autoref{sec:conclusions}.

% ================ end section ====================

\section{Related Work}\label{sec:related-work}

Annotated speech data of sufficient quantity and quality to train end-to-end speech recognizers is scarce for most languages other than English.
Nevertheless, there is demand for high-quality ASR systems for these languages.
Dealing with this issue requires specialized methods.

One such method, known as \emph{transfer learning}, is a machine learning technique for enhancing a model's performance in a data-scarce domain by cross-training on data from other domains or tasks.
There are several kinds of transfer learning.
The predominant one being applied to \ac{ASR} is \emph{heterogeneous transfer learning} \cite{wang_transfer_2015} which involves training a base model on multiple languages (and tasks) simultaneously.
While this achieves some competitive results \cite{chen_mak_2015,knill_gales_2014}, it still requires large amounts of data to yield robust improvements \cite{heigold_multilingual_2013}.

In terms of how much data is needed for effective retraining, a much more promising type of transfer learning is called \emph{model adaptation} \cite{wang_transfer_2015}.
With this technique, we first train a model on one (or more) languages, then retrain all or parts of it on another language which was unseen during the first training round.
The parameters learned from the first language serve as a starting point, similar in effect to pre-training.
\citet{conf/interspeech/VuS13} applied this technique by first learning a \ac{MLP} from multiple languages with relatively abundant data, such as English, and then getting competitive results on languages like Czech and Vietnamese, for which there is not as much data available.

The method presented in this paper differs from \citet{conf/interspeech/VuS13} in that it does not force the representation to be compressed into \emph{bottleneck features} \cite{grezl2008optimizing} and use the result as the output of the pre-trained network.
The idea of freezing only certain layers is another way in which our approach differs.

% ================ end section ====================

\section{Model Architecture}\label{sec:model-architecture}

One of the reasons \citet{DBLP:journals/corr/AmodeiABCCCCCCD15} had to train their network using many GPUs was its complexity.
It uses both convolutional and recurrent units stacked in many layers.
Recently, a much simpler architecture called Wav2Letter has been proposed by \citet{collobert_wav2letter_2016}.
This model does not sacrifice accuracy for faster training.
It relies entirely on its loss function to handle aligning the audio and the transcription sequences while the network itself consists only of convolutional units.

\begin{figure}[tb!]
  \includegraphics[width=\columnwidth]{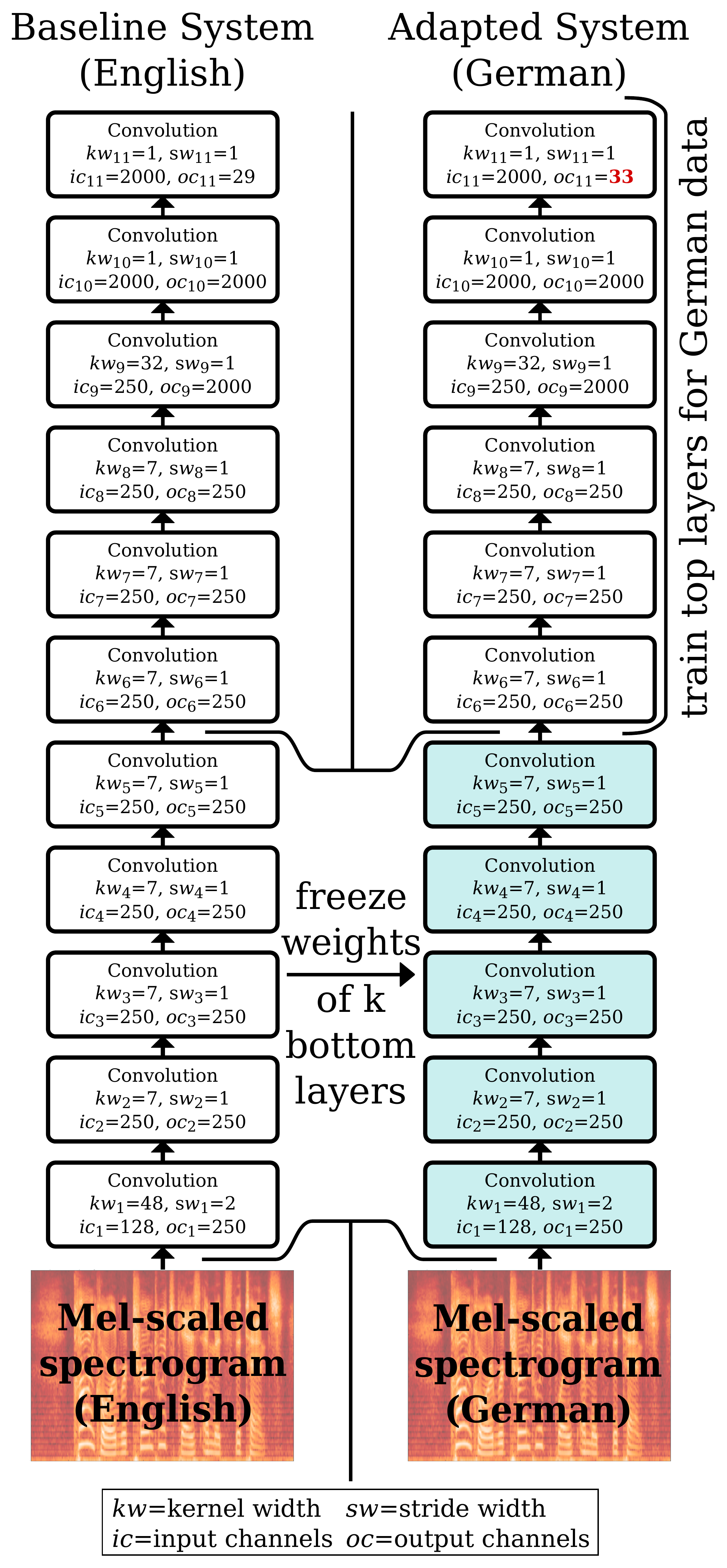}
  \caption{Network architecture adapted from \citet{collobert_wav2letter_2016}.}
  \label{fig:network-architecture}
\end{figure}

The resulting shorter training time and lower hardware requirements make Wav2Letter a solid basis for all of our transfer learning experiments.
Since the general structure of the network is described in the original paper, we only mention what is notable in our adaptation of it in the following.
An overview of their architecture is shown in \autoref{fig:network-architecture}.

\citet{collobert_wav2letter_2016} do not specify the optimizer they used.
We tried several conventional gradient descent optimizers and achieved best convergence with Adam \cite{DBLP:journals/corr/KingmaB14}.
Hyperparameters were slightly adapted from the defaults given by \citet{DBLP:journals/corr/KingmaB14}, that is, we used the learning rate $\alpha = 10^{-4}$, $\beta_{1} = 0.9$, $\beta_{2} = 0.999$ and $\epsilon = 10^{-8}$.
\citet{collobert_wav2letter_2016} note that the choice of activation function for the inner convolution layers does not seem to matter.
We chose rectified linear units as our activation function because they have been shown to work well for acoustic models \cite{maas2013rectifier}.
Weights are initialized Xavier uniformly as introduced by \citet{glorot2010understanding}.

At test time, decoding is performed using a beam search algorithm based on KenLM \cite{Heafield-estimate}.
The decoding procedure follows the TensorFlow implementation based on \cite{DBLP:series/sci/2012-385}.
A beam is scored using two hyperparameters that were derived using a local search optimized to yield the best combined \ac{WER} and \ac{LER} on the LibriSpeech \cite{panayotov_librispeech_2015} validation set.
For the weight of the language model we chose $w_{lm} = 0.8$ and a weight multiplied with the number of vocabulary words in the transcription $w_{valid\_word} = 2.3$.

The \ac{CNN} was implemented in Keras \cite{chollet2015keras}.
The language model and beam search were done in TensorFlow \cite{tensorflow2015-whitepaper} and the introspection in NumPy \cite{DBLP:journals/corr/abs-1102-1523}.
The source code can be found at:
\anonymize{
\href{https://github.com/transfer-learning-asr/transfer-learning-asr}{https://github.com/transfer-learning-asr/transfer-learning-asr}.
}{
Omitted for double-blind review.
}

One of the innovations in \citet{collobert_wav2letter_2016} was the introduction of the \ac{ASG} loss function.
The authors reported it mainly improving the model's throughput with negligible effect on \ac{WER} and \ac{LER} compared to the \ac{CTC} loss introduced by \citet{graves2006connectionist}.
Since there is currently no publicly available implementation of this loss function, we decided to stay with an existing TensorFlow implementation of the \ac{CTC} loss instead.

The English model achieved a \ac{LER} of 13.66\% and \ac{WER} of 43.58\% on the LibriSpeech \cite{panayotov_librispeech_2015} test-clean corpus.
This is worse than the results of \citet{collobert_wav2letter_2016}.
Since the authors of that paper did not publish their source code, we were not able to reproduce their results reliably.
All of our transfer learning experiments are based on this model and for our experiments it is assumed that such a model is already given for the transfer learning task that is to be performed.

% ================ end section ====================

\section{Datasets}\label{sec:datasets}

For training the English model, we used the LibriSpeech corpus \cite{panayotov_librispeech_2015}. 
This dataset consists of about 1000 hours of read speech, sampled at 16 kHz, from the domain of audio books. 
This is the same dataset that was used to train the original Wav2Letter model.

The German models were trained on several corpora taken from the \ac{BAS} \cite{schiel_bas_1998, reichel_bas_2016} as well as the dataset described in \citet{radeck_corpus_2015}, which will be referred to as ``RADECK'' from now on.
Overall, we had a total of 383 hours of training data, which is only slightly more than one third of the English corpus.
Additional quantitative information regarding each corpus, as well as any available references, is given in \autoref{table_corpora_quantitative}.
Information about the kind of recording contained in each corpus is given in \autoref{table_corpora_speechinfo}.
It is also important to point out that some of the corpora pose additional challenges for speech recognition like partially intoxicated people, recordings over telephone, and different dialects.

Each German corpus was split into training and test sets.
We grouped the audio by speakers and used 10\% of the groups for testing.
Therefore, no speaker appears in both training and test set ensuring that results are not due to overfitting to certain speakers.
Exceptions to this procedure are: 
The VM corpora, which were used exclusively for training because obtaining a split based on speakers was not trivial here; 
SC10, which was used only for testing because it consists of recordings of speakers with German as a second language and strong foreign accents with only 5.8 hours in size;
and RADECK, where we used the original splits.

\begin{table*}
\centering
{\footnotesize
%\begin{tabular}{|l|l|p{0.23\textwidth}|l|}
\begin{tabularx}{\textwidth}{l l l l l l l}
\toprule
Name & Size & Number of speakers & S LER & S WER & TL LER & TL WER \\
\midrule
%\hline
ALC \cite{schiel_alc_2012} & 54.54h & 162 & 13.48\% & 32.83\% & 8.23\% & 21.14\% \\
%\hline
HEMPEL \cite{draxler_three-bas-corpora_2002} & 14.21h & 3909 & 34.05\% & 71.74\% & 19.13\% & 46.78\% \\
%\hline
PD1 & 19.36h & 201 & 21.02\% & 34.37\% & 8.32\% & 11.85\% \\
%\hline
PD2 & 4.33h & 16 & 7.60\% & 19.64\% & 1.97\% & 5.96\% \\
%\hline
RVG-J \cite{draxler_three-bas-corpora_2002} & 46.28h & 182 & 17.43\% & 39.87\% & 10.85\% & 24.92\% \\
%\hline
SC10 & 5.80h & 70 & 25.62\% & 78.82\% & 17.59\% & 57.84\% \\
%\hline
VM1 \cite{wahlster_verbmobil_1993} & 32.40h & 654 & - & - & - & - \\
%\hline
VM2 \cite{wahlster_verbmobil_1993} & 43.90h & 214 & - & - & - & - \\
%\hline
ZIPTEL \cite{draxler_three-bas-corpora_2002} & 12.96h & 1957 & 22.87\% & 62.27\% & 15.07\% & 46.25\% \\
%\hline
RADECK \cite{radeck_corpus_2015} & 181.96h & 180 & 27.83\% & 65.13\% & 20.83\% & 56.17\% \\
\midrule
All corpora & 415.7h & 7545 & 22.78\% & 58.36\% & 15.05\% & 42.49\% \\
\bottomrule
\end{tabularx}
}
\caption{Quantitative information on the corpora used to train the German model.
References to individual corpora are given where available.
Size and number of speakers refer only to the subsets we used (including training and test sets).
Test set \ac{LER} and \ac{WER} are reported for the best transfer learning (TL) model and the model from scratch (S) after 103h of training.}
\label{table_corpora_quantitative}
\end{table*}

\begin{table*}
\centering
{\footnotesize
\begin{tabularx}{\textwidth}{l l X X}
\toprule
Name & Speech Type & Topic & Idiosyncrasies \\
\midrule
ALC & read, spontaneous & car control commands, tongue twisters, answering questions & partially recorded in running car; speakers partially intoxicated \\
%\hline
HEMPEL & spontaneous & answer: What did you do in the last hour? & recorded over telephone \\
%\hline
PD1 & read & phonetically balanced sentences, two stories: ``Buttergeschichte'' and ``Nordwind und Sonne'' & recordings were repeated until error-free \\
%\hline
PD2 & read & sentences from a train query task & recordings were repeated until error-free \\
%\hline
RVG-J & read, spontaneous & numbers, phonetically balanced sentences, free-form responses to questions & speakers are adolescents mostly between the ages 13--15 \\
%\hline
SC10 & read, spontaneous & phonetically balanced sentences, numbers, ``Nordwind und Sonne'', free dialogue, free retelling of ``Der Enkel und der Grossvater'' & multi-language corpus; only German data was used \\
%\hline
VM1 & spontaneous & dialogues for appointment scheduling & multi-language corpus; only German data was used \\
%\hline
VM2 & spontaneous & dialogues for appointment scheduling, travel planning and leisure time planning & multi-language corpus; only German data was used \\
%\hline
ZIPTEL & read & street names, ZIP codes, telephone numbers, city names & recorded over telephone \\
%\hline
RADECK & read, semi-spontaneous & Wikipedia, European Parliament transcriptions, commands for command-and-control settings & contains six microphone recordings of each speech signal \\
\bottomrule
\end{tabularx}
}
\caption{Information on the kind of speech data contained in each corpus.}
\label{table_corpora_speechinfo}
\end{table*}

We also rely on text corpora for the KenLM decoding step.
For the English corpus \cite{panayotov_librispeech_2015}, the provided 4-gram model based on all training transcriptions was used like in the original Wav2Letter implementation.
For the German corpus, our n-gram model came from a preprocessed version of the German Wikipedia, the European Parliament Proceedings Parallel Corpus\footnote{\href{https://github.com/tudarmstadt-lt/kaldi-tuda-de/}{https://github.com/tudarmstadt-lt/kaldi-tuda-de/}}, and all the training transcriptions.
Validation and test sets were carefully excluded.

% =============== end subsection ==================

\subsection{Preprocessing}

Since the English model was trained on data with a sampling rate of 16 kHz, the German speech data needed to be brought into the same format so that the convolutional filters could operate on the same timescale.
To this end, all data was resampled to 16 kHz.
Preprocessing was done using librosa \cite{mcfee_librosa_2015} and
consisted of applying a \ac{STFT} to obtain power level spectrum features from the raw audio as described in \citet{collobert_wav2letter_2016}.
After that, spectrum features were mel-scaled and then directly fed into the \ac{CNN}.
Originally, the parameters were set to window length $w = 25\mathrm{ms}$, stride $s = 10\mathrm{ms}$ and number of components $n = 257$.
We adapted the window length to $w_{new} = 32\mathrm{ms}$ which equals a Fourier transform window of $512$ samples, in order to follow the convention of using power-of-two window sizes.
The stride was set to $s_{new} = 8\mathrm{ms}$ in order to achieve 75\% overlap of successive frames.
%It also turns out 
We observed
that $n = 257$ results in many of the components being $0$ due to the limited window length.
We therefore decreased the parameter to $n_{new} = 128$.
After the generation of the spectrograms, we normalized them to mean 0 and standard deviation 1 per input sequence.

Any individual recordings in the German corpora longer than 35 seconds were removed due to GPU memory limitations. 
This could have been solved instead by splitting audio files using their word alignments where provided (and their corresponding transcriptions), but we chose not to do so since the loss of data incurred by simply ignoring overly long files was negligible.
Corpora sizes given in \autoref{table_corpora_quantitative} are after removal of said sequences.
We excluded 1046 invalid samples in the RADECK corpus due to truncated audio as well as 569 samples with empty transcriptions.

% =============== end subsection ==================

\section{Experiments}\label{sec:experiments}

Given the English model, we froze $k$ of the lower layers and trained all $11 - k$ layers above with the German corpora.
This means the gradient was only calculated for the weights of those $11 - k$ layers and gradient descent was then applied to update those as usual.
The process of freezing $k$ layers is visualized in \autoref{fig:network-architecture}.
The transfer training was performed based on both the original weights as well as a new random initialization for comparison.
Except for changing the training data, the German corpora introduce four new class labels \emph{\"a\"o\"u\ss{}} in addition to the original 28 labels.
We set the initial weights and biases of the final softmax layer for these labels to zero.
Additionally, as a baseline for the performance of a Wav2Letter based German \ac{ASR}, we trained one model from scratch on all German training corpora.
For all experiments we used a batch size of $64$, both during training as well as evaluation.

% =============== end subsection ==================

\section{Results and Discussion}\label{sec:results}

As initially hypothesized, transfer learning could give us three benefits:
Reduced computing time, lower GPU memory requirements and a smaller required amount of German speech data.
In addition to that, we may find structural similarities between languages for the \ac{ASR} task.
In the subsequent sections, we will first report general observations,
evaluate each hypothesis based on the performed experiments and then analyze the learned weights using introspection techniques.
We report overall test scores and scores on each test set in the form of \acp{WER} and \acp{LER}.
Finally, we discuss the language specific assumptions that were required for the experiments and how transfer learning may perform on other languages.

\subsection{Retaining or reinitializing weights?}

\begin{figure}[th]
  \includegraphics[width=\columnwidth]{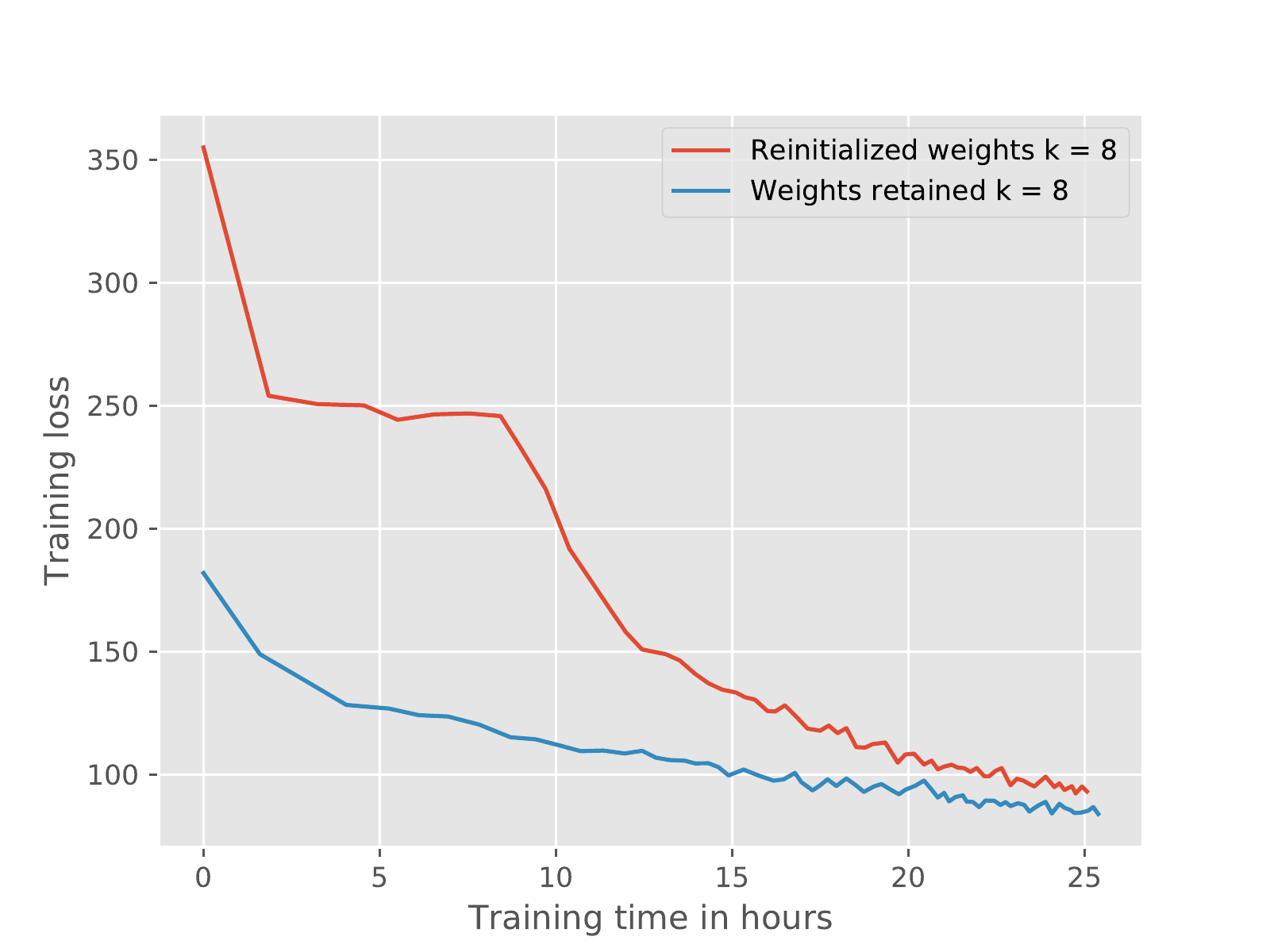}
  \caption{Comparison of learning curves for 25 hours of training with either reinitialized or retained weights. In both cases $k = 8$ layers were frozen.
  }
  \label{fig:convergence-reinit}
\end{figure}

When the transfer learning training is performed, one could either continue training on the existing weights or reinitialize them.
Reusing existing weights might lead to \ac{SGD} being stuck in a local minimum, reinitializing may take longer to converge.
For $k = 8$ we compared the speed of training for both methods.
As it can be seen in \autoref{fig:convergence-reinit}, using existing weights is much faster without a decrease in quality.

\subsection{Reduced computing time}

\begin{figure}[th]
  \includegraphics[width=\columnwidth]{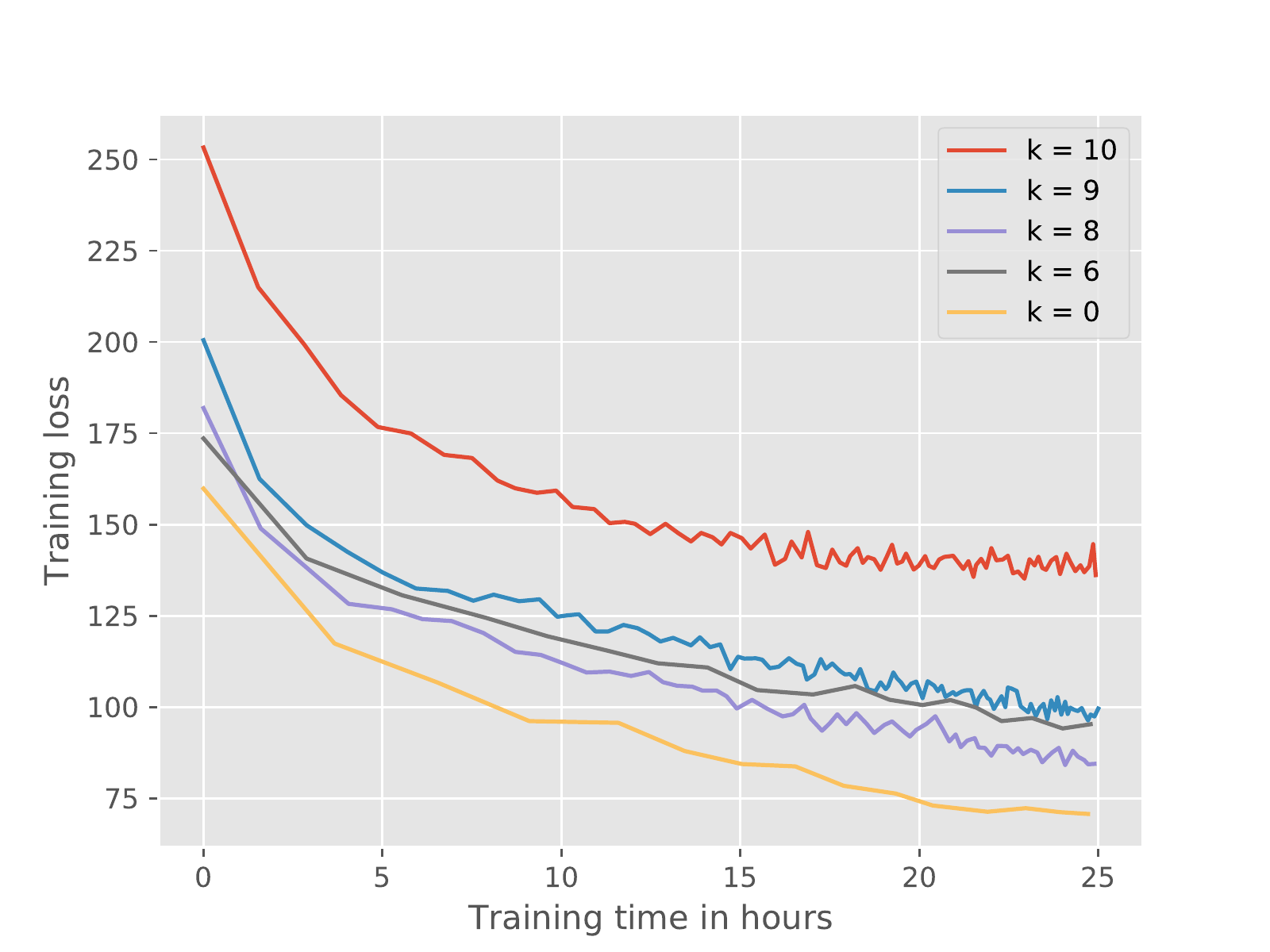}
  \caption{Learning curves for 25 hours of training with different numbers $k$ of frozen layers.
Note that due to the decreased time to process a batch (cf.~\autoref{fig:time-per-step}),
training models with higher $k$ (more frozen layers) allows to iterate over the training data more often in the same amount of time.
But eventually, this does not help to beat the model with $k=0$ which is trained with the fewest dataset iterations but still at any time achieves the lowest loss.
  }
  \label{fig:loss-different-k}
\end{figure}

Given that languages share common features in their pronunciation,
lower layers should contain common features that can be reused when transferring the model to a different language.
Therefore, we subsequently froze $k$ layers of the original English model,
choosing a different $k$ in each experiment.
Our experiments showed that this assumption is indeed true,
it is sufficient to adjust only at least two layers for achieving training losses below 100 after 25 hours.
The loss curve for different $k$ can be seen in \autoref{fig:loss-different-k}.

\begin{figure}[th]
  \includegraphics[width=\columnwidth]{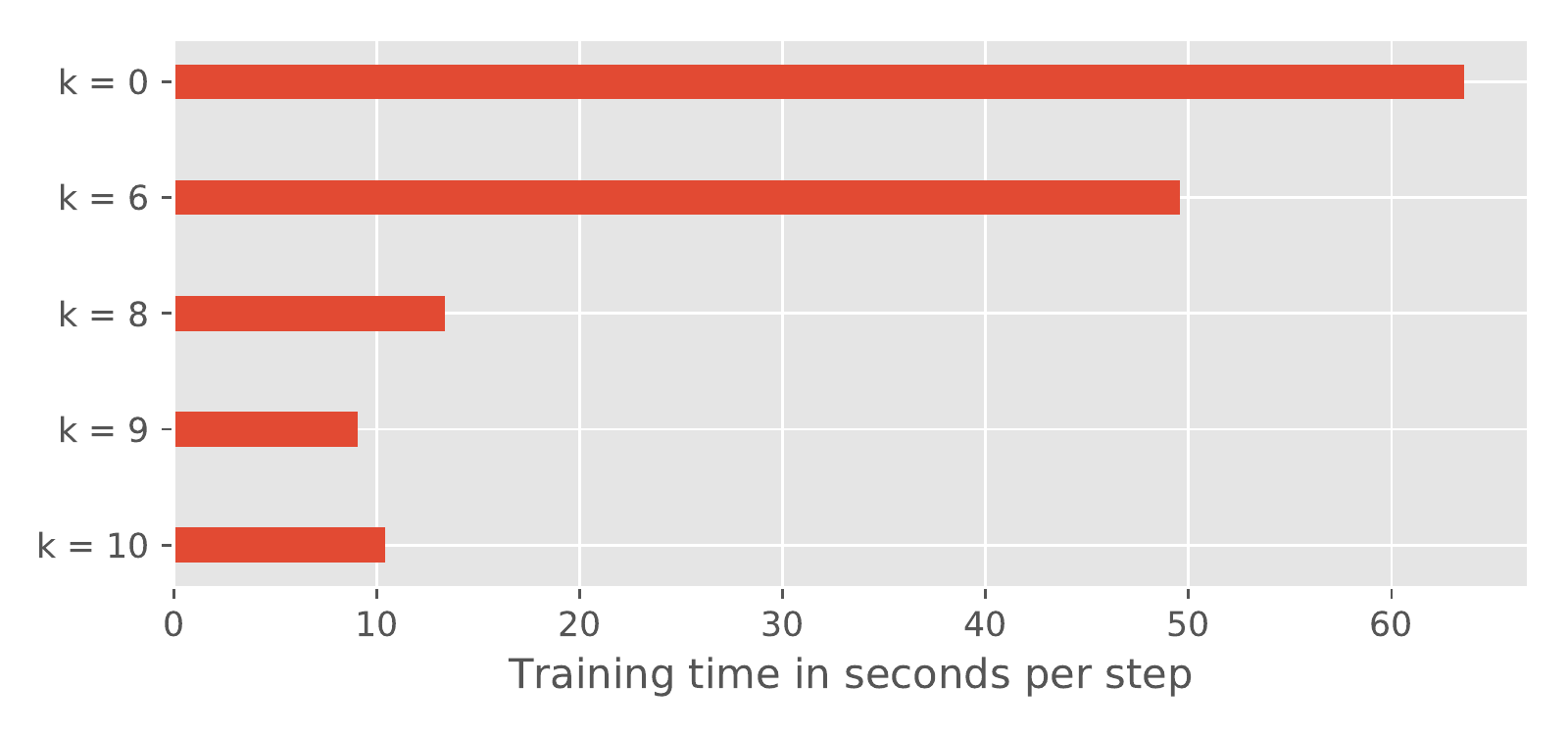}
  \caption{The more layers we freeze, the faster one batch of $64$ is trained. Measured over 25h of training each.}
  \label{fig:time-per-step}
\end{figure}

For bigger $k$ we need to backpropagate through fewer layers, therefore training time per step (training one batch) decreases almost monotonically with $k$ in \autoref{fig:time-per-step}.
Despite that boost in training time, experiments show that loss is almost always smaller at any given point in time for smaller $k$.
In \autoref{fig:loss-different-k} this manifests in $k = 0$ always having the smallest training loss.
We conclude that in terms of achieving small loss, there is no reason to favor big values for $k$, freezing layers is not necessary.

\begin{figure}[th]
  \includegraphics[width=\columnwidth]{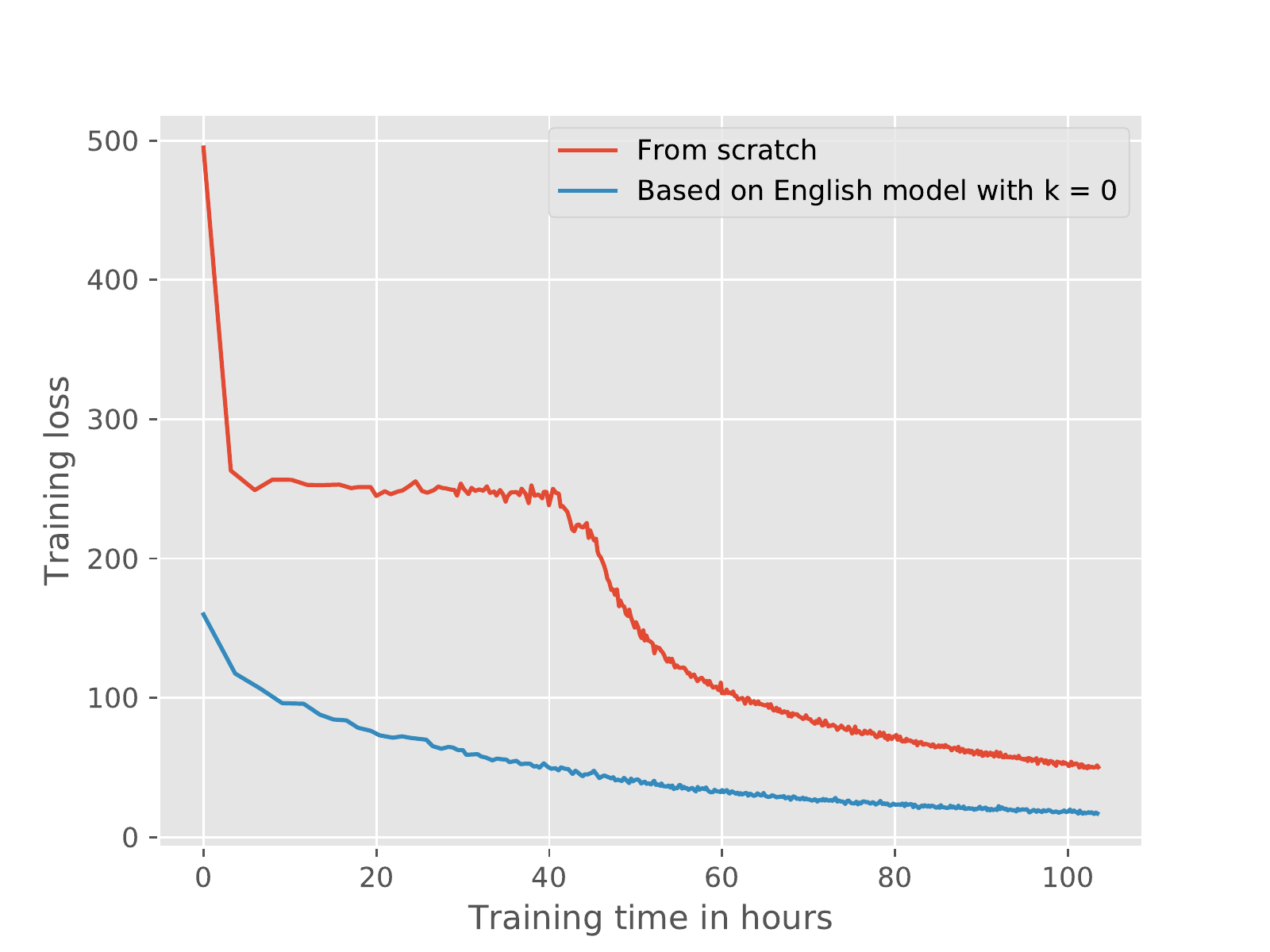}
  \caption{Applying transfer learning by using the weights from the English model leads to small losses more quickly than training from scratch.}
  \label{fig:scratch-vs-retained}
\end{figure}

When we compare the best transfer learning model with $k = 0$ with a German model trained from scratch in \autoref{fig:scratch-vs-retained},
we are able to see huge improvements in terms of computing time required for achieving the same loss.
We conclude that a good weight starting configuration from another language's \ac{ASR} is beneficial.

\subsection{Lower GPU memory requirements}

Not only does it matter how long training takes with given resources,
many researchers may also have only limited GPU memory at disposal.
All of our experiments were performed on a single GeForce GTX Titan X graphics card,
but the more layers $k$ we freeze, the fewer layers we need to backpropagate through.
Therefore, memory requirements for the GPU are lower.
For a batch size of 64, forward propagation takes less than 3 GB of memory,
while training the whole network requires more than 10.4 GB.
In contrast to that, freezing 8 layers already enables training with less than 5.5 GB of GPU memory.

\subsection{Little German speech data required}

\begin{figure}[th]
  \includegraphics[width=\columnwidth]{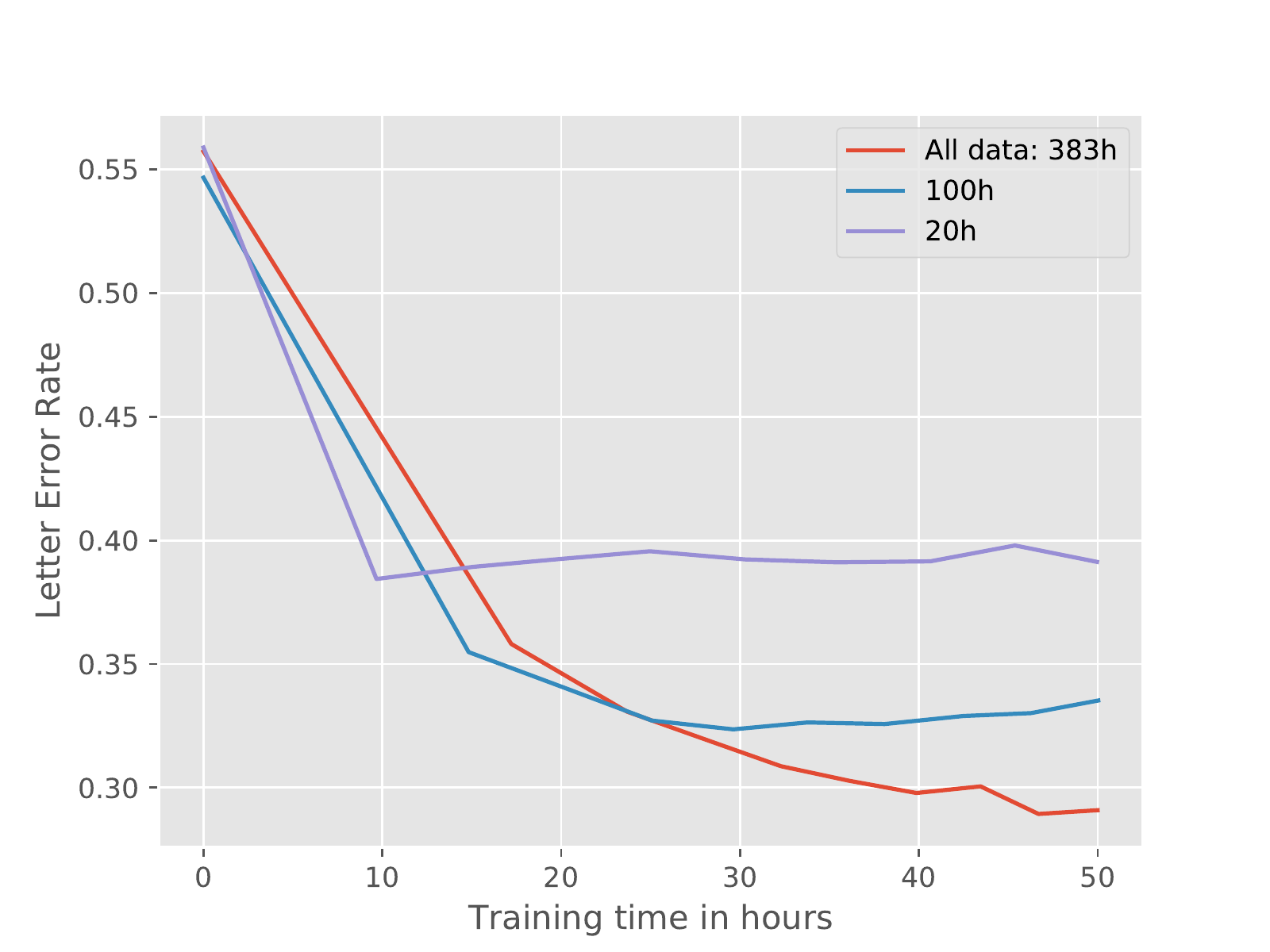}
  \caption{\ac{LER} as a mean over all test samples for different training set sizes with k = 8 for all experiments}
  \label{fig:less-data}
\end{figure}

We hypothesized that little training data may be required for the transfer learning task.
Additionally to using the whole 383 hours of data we had available, we also tried an even more scarce variant.
In order to prevent overfitting, we used a transfer learning model with $k = 8$ for our experiments.
As it can be seen in \autoref{fig:less-data}, for a model with $k = 8$ that was trained for 25 hours, the \ac{LER} using 100 hours of audio is almost equal to using the complete training data.
Longer training causes overfitting.
When using just 20 hours of training data this problem occurs even earlier.
We can conclude that even though training for just 25 hours works well with only 100 hours of audio, beyond that overfitting appears nevertheless.

\subsection{Model Introspection}

\begin{figure*}[th!]
  \includegraphics[width=\linewidth]{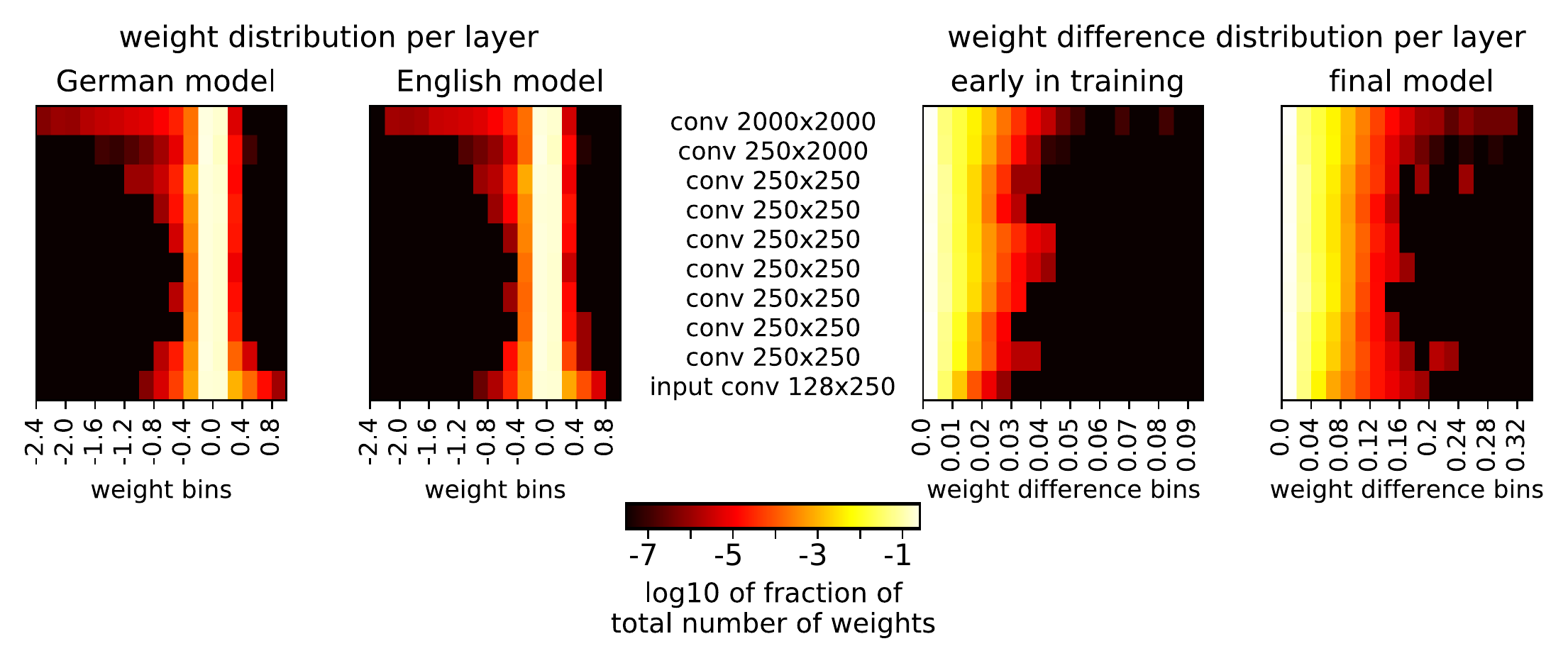}
  \caption{Weight distributions of the German and English model (left) and weight difference distributions both in an early stage and for the final model (right).}
  \label{fig:weight-dist-diff}
\end{figure*}

When applying transfer learning, it is of interest how much the model needs to be adapted and which portions of the model are shared between different languages.
To get insights into those differences, we compared the learned parameters both between the English model and adapted German model (for $k=0$) as well as between different points in time during training.
Since the output layers of both models do not use the same number of output features, we excluded this layer from the comparison.
First, we investigated the distribution of weights and corresponding changes between the English and adapted model, visualized on the left side of \autoref{fig:weight-dist-diff}.
The plot shows the fraction of weights in a layer lying in the respective range of values.
Because most of the weights are between -0.2 and 0.2 (in just 2 bins), we used a log$_{10}$-scale for the fraction of weights in each bin.
We observed that the weights of highest absolute values are in the input and topmost layer.
This indicates that the transformations in the middle layers are smaller than in the outer ones.
Moreover, the weights of each layer are distributed with a mean value close to zero and very small variance.
Due to the similar distributions, it is reasonable to compare the weights and their differences in the following.
Between both models, there are only minor changes in the weight distributions, which supports the assumption that transfer learning is performing well because the English model is a suitable model for being adapted to German.

Since the adaptation to German is not explainable based on the distributions, we further investigated the differences between the individual weights.
Therefore, we determined the absolute distance between weights as shown in \autoref{fig:weight-dist-diff} on the right side.
In the plot, we visualize the distribution of weight changes.
We observed only small changes, therefore a log$_{10}$-scale is used again.
\autoref{fig:weight-dist-diff} on the right side shows this analysis for the transfer learning model early in training as well as the final model after four days.
In the early phase, weights had only been adapted little with a maximum difference of 0.1, while the final model weights changed up to 0.36.
Additionally, we observed that the weights changed more in the middle and top layers earlier, but with progressing training the input layer experiences more changes.
This higher variability in the outer layers can both be observed in the weights of each individual model as well as in their differences.
That is an indication that the model needs to alter the outer layers more than the inner ones in order to adapt to a particular language. 

\begin{figure}[h!]
  \includegraphics[width=\columnwidth]{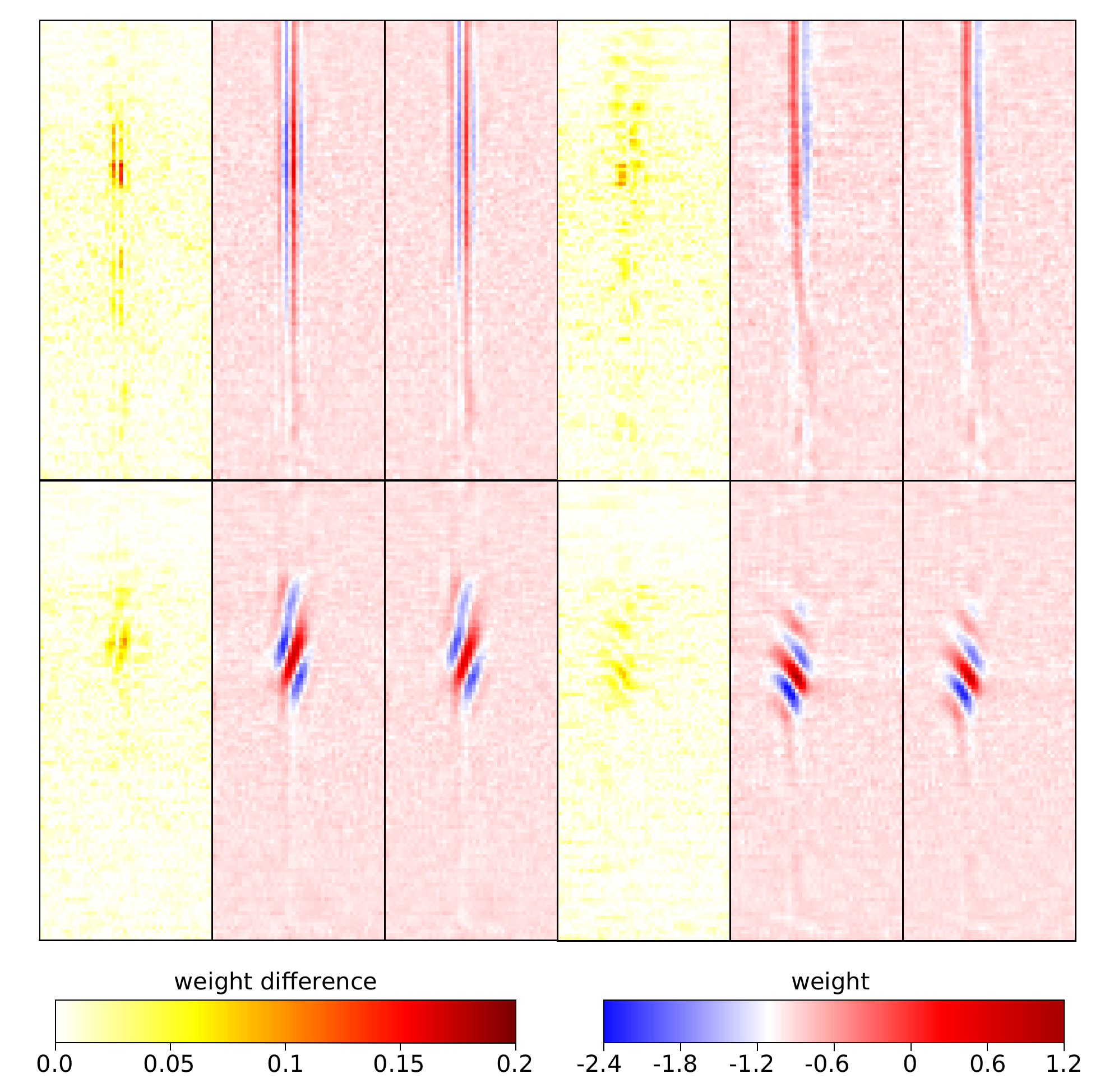}
  \caption{Differences in specific filters of the input layer. Neurons were chosen based on particular patterns. Each triplet of images shows the weight differences and the corresponding weights in the German and English model (from left to right).}
  \label{fig:filter-introspection}
\end{figure}

Finally, we looked into the changes of individual filters.
Due to the large number of neurons, we provide the complete set of filters from all layers only in the supplement.\footnote{supplements:
\href{https://doi.org/10.6084/m9.figshare.5048965}{https://doi.org/10.6084/m9.figshare.5048965}}
We present our findings for a selected set of neurons of the input layer that showed well-interpretable patterns.
The weights of those filters and their differences between the English and German model are shown in \autoref{fig:filter-introspection}.
The top row shows neurons that can be interpreted as detectors for short percussive sounds (e.g.~t or k) and the end of high pitched noise (e.g.~s).
The bottom neurons might detect rising or falling pitch in vowels.
Of these four filters, the upper left differs most between English and German with a maximum difference of 0.15.
This supports that it is detecting percussive sounds as German language has considerably stronger pronunciation of corresponding letters than English.
On the other hand, the bottom row filters experienced less change (both $<0.1$ maximum difference).
This supports them being related to vocal detection since there are few differences in pronunciation between English and German speakers.

\subsection{Overall test set accuracy}

All test set \acp{LER} and \acp{WER} scores are consistent with the differences of loss in the performed experiments.
After 103 hours of training, the best transfer learning model is therefore $k = 0$ with a \ac{LER} of $15.05\%$ and \ac{WER} of $42.49\%$ as the mean over all test samples.
The model that has been trained from scratch for the same amount of time achieves a \ac{LER} of $22.78\%$ and \ac{WER} of $58.36\%$.
\autoref{table_corpora_quantitative} gives details about the accuracy on each test set.

Some very high \acp{WER} are due to heavy German dialect that is particularly problematic with numbers, e.g.
\begin{framed}
\noindent
Expected:  ``sechsundneunzig'' \\
Predicted: ``sechs un nmeunsche'' \\
LER 47\%, WER 300\%, loss: 43.15
\end{framed}
This shows, that there is both room for improvement in terms of word compounds as well as \ac{ASR} of different dialects where data is even more scarce.

\subsection{Accuracy boost through language model decoding}

The original Wav2Letter network did not report on improvements in \ac{LER} and \ac{WER} due to the KenLM integration.
In \autoref{fig:kenlm-boost} We compared decoding performed through KenLM scored beam search with a greedy decoding on the German corpora.

\begin{table}[h]
\begin{tabular}{l l l}
  \toprule
  & LER & WER \\
  \midrule
  with LM & 15.05\% & 42.49\% \\
  without LM & 16.77\% & 56.14\% \\
  \bottomrule
\end{tabular}
\centering
\caption{Comparing \ac{LER} and \ac{WER} with and without KenLM based on model with $k = 0$}
\label{fig:kenlm-boost}
\end{table}

\subsection{Transfer learning for other languages}
In our speech recognizer, the lower layers of the network learn phonological features whereas the higher (deeper) ones map these features onto graphemes. 
Thus for \ac{ASR} these two types of features clearly matter the most.
German and English have many phonemes and graphemes in common.
The apparent success of our transfer learning approach was greatly facilitated by these similarities.
Not all languages share as much in terms of these features.
We anticipate that our approach will be less effective for such pairs.
This means we expect the adaptation to a less similar language to require more data and training time.
We further suspect that differences in grapheme inventories cause other effects than differences in phonemes.
This is because only the mapping of phonological features to graphemes has to be adapted for a different orthography.
In contrast, differences in phoneme inventories require more changes in features learned at lower layers of the network.
Moreover, there could be differences in the importance of specific features.
For instance, having vowels in common is potentially more important for transfer learning than sharing many consonants, because vowels experience higher variability in pronunciation.
At the same time very drastic differences in orthography could probably trigger a stronger change of weights in lower network layers.
We expect our transfer learning approach to encounter strong difficulties sharing knowledge between English and a logographic language like Mandarin Chinese.
Despite those difficulties, using weights from a pre-trained \ac{ASR}-network is a more reasonable initialization than random weights.
This is because very basic audio features are shared between all languages.
Therefore even for more different language pairs, we expect transfer learning to decrease the necessary amount of training data and time.

% ================ end section ====================

\section{Conclusions}\label{sec:conclusions}

We were able to show that transfer learning using model adaptation can improve the speed of learning when only 383 hours of training data are available.
Given an English model, we trained a German model that outperforms the German baseline model trained from scratch in the same amount of training time.
Thus, with little time, our approach allows training better models.
We showed that the English model's weights are a good starting configuration and allow the transfer learning model to reach smaller training losses in comparison to a weight reinitialization.
When less GPU memory is available, freezing the lower 8 layers allows to train batches of 64 with less than 5.5 GB instead of more than 10.4 GB while still performing similar after 25 hours of training.
Model introspection determined that lower and upper layers, in contrast to the layers in the center, need to change more thoroughly in order to adapt to the new language.

We identified several interesting directions for future work.
Test accuracy showed that word compounds can be challenging and dialects pose difficulties when little training data is available.
GPU memory consumption could be further reduced by caching the representation that needs only forward propagation.
An open source version of the \ac{ASG} loss would enable faster training.
Finally, future research should investigate how well this transfer learning approach generalizes by applying it to more distinct languages.

% ================ end section ====================

\section*{Acknowledgments}
%The acknowledgments should go immediately before the references.  Do
%not number the acknowledgments section. Do not include this section
%when submitting your paper for review.
\anonymize{
This research was supported by the donation of a GeForce GTX Titan X graphics card from the NVIDIA Corporation.
}{
Omitted for double-blind review.
}

%\clearpage % remove this later! Just to see how much space we got

%\bibliography{acl2017}
\bibliography{acl2017}
\bibliographystyle{acl_natbib}

%\appendix
%
%\section{Supplemental Material}
%\label{sec:supplemental}

\end{document}